\documentclass[10pt,a4paper,conference]{IEEEtran}

\usepackage{graphicx}
\usepackage{inputenc}
\usepackage{color}
\usepackage{hyperref}
\hypersetup{
    colorlinks=true,
    linkcolor=black,
    filecolor=magenta,      
    urlcolor=cyan,
    citecolor=black,
}

\ifCLASSINFOpdf
\else
\fi
\begin{document}
%
\title{Future Urban Scenes Generation Through Vehicles Synthesis}

\author{\IEEEauthorblockN{Alessandro Simoni, Luca Bergamini, Andrea Palazzi, Simone Calderara, Rita Cucchiara}
\IEEEauthorblockA{University of Modena and Reggio Emilia, Modena, Italy \\
\{alessandro.simoni, luca.bergamini24, andrea.palazzi, simone.calderara, rita.cucchiara\}@unimore.it}}


%


\maketitle

\begin{abstract}
In this work we propose a deep learning pipeline to predict the visual future appearance of an urban scene. Despite recent advances, generating the entire scene in an end-to-end fashion is still far from being achieved. Instead, here we follow a two stages approach, where interpretable information is included in the loop and each actor is modelled independently. We leverage a per-object \textit{novel view synthesis} paradigm; i.e. generating a synthetic representation of an object undergoing a geometrical roto-translation in the 3D space. Our model can be easily conditioned with constraints (e.g. input trajectories) provided by state-of-the-art tracking methods or by the user itself. This allows us to generate a set of diverse realistic futures starting from the same input in a \textit{multi-modal} fashion. We visually and quantitatively show the superiority of this approach over traditional end-to-end scene-generation methods on CityFlow, a challenging real world dataset.
\end{abstract}
\section{Introduction}
In the near future, smart interconnected cities will become reality in various countries worldwide. In this scenario, vehicles -- both autonomous and not -- will play a fundamental role thanks to key technologies developed to connect them (e.g. 5G) and advanced sensors (e.g. lidars, radars) enabling a deeper understanding of the scene.
Explainability is expected to be a mandatory requirement to ensure the safeness of all other actors (including pedestrians, cyclists, \dots). However, the current approach to autonomous driving related tasks is still end-to-end, which greatly obscures the learned knowledge. Despite that, recent works~\cite{bansal2018chauffeurnet,hong2019rules} have moved from this framework -- where raw inputs are transformed into the final outputs/decision -- to a more interpretable one, where an intermediate high level representation is employed. Those representations can be easily understood by human operators and provide an effective parallelism between human and autonomous decision taking.

\begin{figure}
    \centering
    \includegraphics[width=0.95\columnwidth]{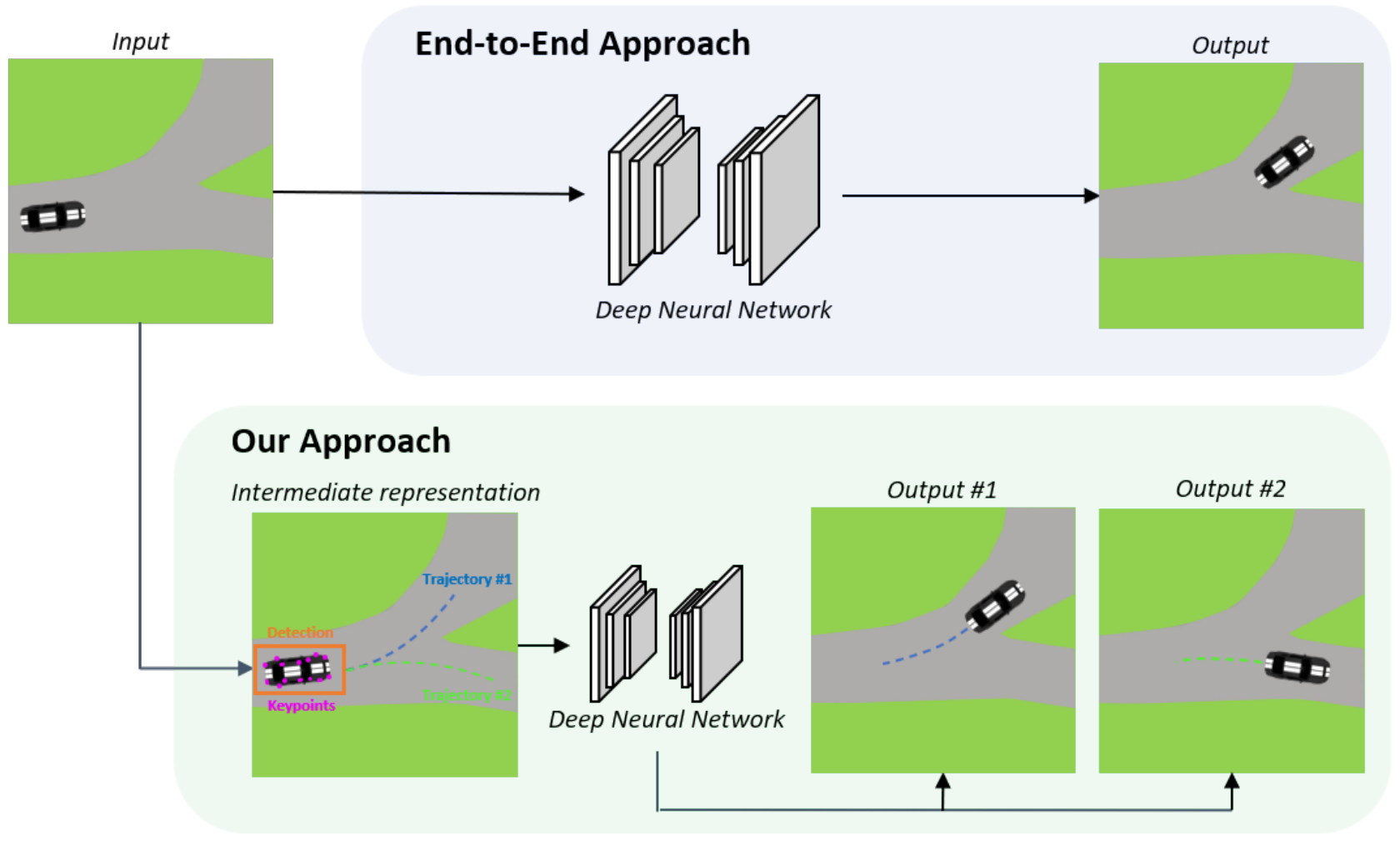}
    \caption{The difference between a black box end-to-end method and our approach, which exploits intermediate interpretable information to synthesise each vehicle individually.}
    \label{overviewfig}
\end{figure}

In this work we take a step forward and present a pipeline where the final output is produced by applying a sequence of operations that mimic those of a human operator for a specific task related to the autonomous driving. In particular, we focus on generating realistic visual futures for urban scenes where vehicles are the main actors. In more details, starting from one or multiple RGB frames, the final output is a clip of images where all the actors in the scene move following a plausible path. In doing so, as depicted in Fig.~\ref{overviewfig}, we rely heavily on information a user can easily understand, such as bounding boxes, trajectories and keypoints. 
Moreover, we wish to easily condition the output on that information; in particular, given a set of trajectories for the same vehicle (either by a state-of-the-art trajectory predictor or a user's input), we would like to generate a set of realistic visual representation of the vehicle following these trajectories. In the following, we focus on vehicles only and leave the analysis of other agents as future work.

It is worth noting how the same task can be tackled as an image-to-image problem, where a deep neural network transforms past frame/s into future ones, as depicted in figure~\ref{overviewfig}.
While many end-to-end methods~\cite{goodfellow2014generative, isola2017image, zhu2017unpaired, wang2018high} can in fact be applied to visual scene generation, they all share some intrinsic drawbacks. In particular: \textit{i)} because they start from raw inputs (i.e. RGB images), 
it is not always clear which is the best way to include user's or geometric constraints; \textit{ii)} despite recent advances in model explanation~\cite{springenberg2014striving,selvaraju2017grad}, end-to-end methods are difficult to investigate either before or after critical faults, which is required for critical applications; \textit{iii)} 
these methods do not focus on the actors but instead transform the entire image, including the static background: this wastes computational time while limiting the maximum resolution that these methods can handle; and \textit{iv)} they can hardly leverage any established state-of-the-art method for additional information, such as vehicle detection or trajectory prediction. 

Contrarily, we frame the task as a two stages pipeline where only vehicles are individually transformed. First, we extract interpretable information from raw RGB frames, including bounding boxes and trajectory estimations. Second, we employ it to produce visual intermediate inputs. Finally, these inputs condition a deep convolutional neural network~\cite{esser2018variational,palazzi2019warp} to generate the final visual appearance of the vehicle in the future.
We argue that this approach is closer to the human way of thinking and, as such, better suits a human-vehicle interactions setting. Similarly to what~\cite{bansal2018chauffeurnet,hong2019rules} devise for autonomous planning, our method offers an interpretable intermediate representation a user can naturally understand and interact with.
Finally, the input resolution does not represent a limit in our proposal. In fact, as only individual vehicles are processed in our pipeline, the input resolution is typically much lower than the full frame one. 

To sum up, we:
\begin{itemize}
    \item Provide a novel pipeline that leverage interpretable information to produce a deterministic visual future grounded on those constraints;
    \item prove that our method is not limited to a uni-modal output, but allows to generate \textit{"alternative futures"} by acting on the intermediate constraints;
    \item show how this approach outperforms end-to-end image-to-image translation solutions both visually and quantitatively.
\end{itemize}

\section{Related works}\label{sec:related}
We first introduce here the current state of the art for image-to-image translation, where one or multiple images are produced starting from a single or a set of input frames. These methods focus on the entire scene, without acknowledging specific elements in the scene.
We then focus on approaches based on view synthesis, where the attention is placed instead on the actor solely, with the aim of producing a novel view of it from a different point of view.

\textbf{Image to Image Translation.} Generative Adversarial Networks (GANs)~\cite{goodfellow2014generative, denton2015deep, mathieu2016disentangling, salimans2016improved, arjovsky2017WassersteinGAN} have been widely used to perform image transformations with impressive results. They exploit an \textit{adversarial loss} to constrain generated images to be as similar as possible to the real ones. This supervision signal generates sharper results when compared with standard maximum estimation based losses, and allows these methods to be employed for image generation and editing tasks associated with computer graphics.

Recent works~\cite{mirza2014conditional, isola2017image, zhu2017unpaired} prove that GANs can help solving conditioned image generation, where the network yields an output image conditioned on an observed input image \textit{x} and an optional random noise \textit{z}. This can be applied for example to transform a segmentation map into image, or a picture taken at day time into one acquired at night time as presented by \cite{isola2017image}. 

Wang et al.~\cite{wang2018high} propose a framework able to synthesise high-resolution images (\textit{pix2pixHD}), while Zhu et al.~\cite{zhu2017unpaired} define the concept of \textit{cycle consistency loss} to supervise GANs training without the need for coupled data; their goal is to define a function \textit{G}, which maps from the first domain to the second, and a function \textit{F}, which performs the opposite. The two domains are bound to be consistent with each other at training time.

With the aim of predicting multiple frames, several works~\cite{lotter2016deep,villegas2019high,qi20193d,kwon2019predicting} extend the image-to-image approach by including time. 
Authors of PredNet~\cite{lotter2016deep} propose a network based on Long Short Term Memory (LSTM)~\cite{hochreiter1997long} combined with convolutional operations to extract features from input images. In~\cite{villegas2019high}, an LSTM based network is trained without any additional information (e.g. optical flow. segmentation masks,\dots) by leveraging the concept of "network capacity maximisation". Qi et al.~\cite{qi20193d} decompose the task of video prediction into ego and foreground motion, and leverage RGBD input for 3D decomposition. Finally, authors from~\cite{kwon2019predicting} address the issue of low quality predictions for distant future by training a network to predict both future and past frames and by enforcing retrospective consistency.

\textbf{View synthesis.} In the last few years, deep generative models have been applied also to novel view generation, i.e. synthesising the aspect of an object from different points of view. Many works~\cite{ma2017pose, zhao2018multi} achieve impressive results on human pose appearance generation. Among them, VUnet~\cite{esser2018variational} is based on a U-Net architecture~\cite{ronneberger2015u} which combines a GAN mapping an estimated shape $y$ to the target image $x$ with a Variatonal AutoEncoder (VAE)~\cite{kingma2013auto} that conditions on the appearance $z$. This network aims to find the maximum a posteriori $p(x | y, z)$, i.e. the best object synthesis conditioned on both appearance and shape constraints.
Yang et al.~\cite{yang2015weakly} propose a recurrent encoder-decoder network to learn how to generate different views of the same object from different rotations. The initial object appearance is encoded into a fixed low dimensional representation, while sequential transformations act on a separate embedding. Finally, the decoder combines both vectors and yields the final prediction.

\begin{figure*}[t]
    \centering
    \includegraphics[width=0.85\textwidth]{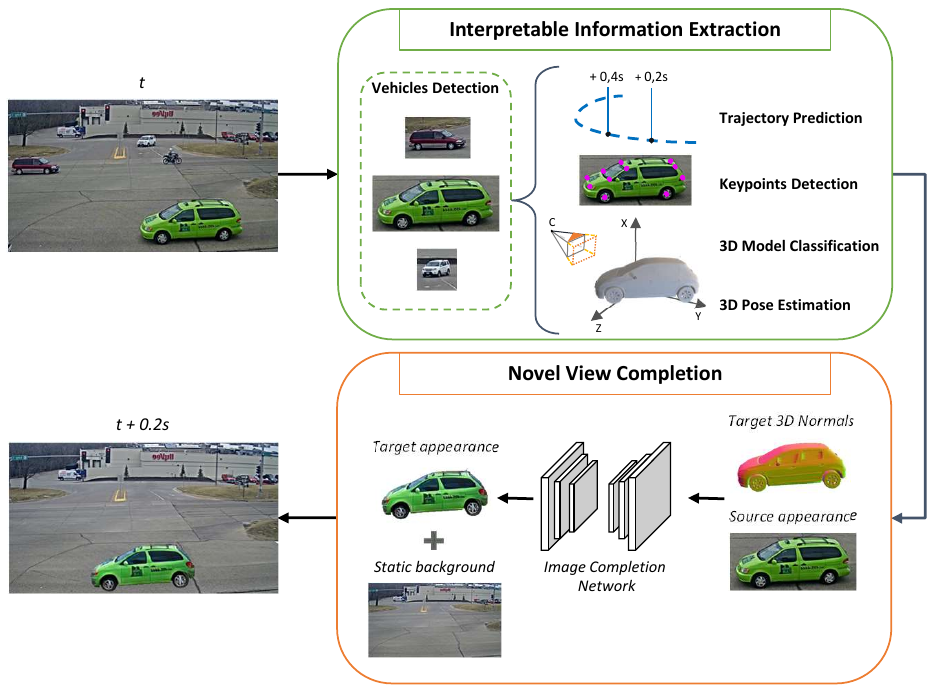}
    \caption{Our model pipeline composed by two stages: (i) \textit{interpretable information extraction} for each vehicle (detection \& tracking), and (ii) \textit{novel view completion} process exploiting the 3D projected rendering of the object (\textbf{target 3D normals}) and its appearance from the cropped image (\textbf{source appearance}).}
    \label{modelfig}
\end{figure*}

In the automotive field, Tatarchenko et al.~\cite{tatarchenko2016multi} train a CNN to estimate the appearance and the depth map of an object after a viewpoint transformation. The transformation is encoded as azimuth-elevation-radius and is concatenated to the appearance embedding after being forwarded through fully connected layers. By combining multiple predicted depth maps, their approach can generate reconstructed 3D models from a single RGB image. Again, Zhou et al.~\cite{zhou2016view} extract appearance flow information to guide pixels locations after an arbitrary rotation. Their model leverages a spatial transformer~\cite{jaderberg2015spatial} to output a grid of translation coefficients. Contrarily, Warp\&Learn~\cite{palazzi2019warp} first extracts 2D semantic patches from the vehicle input image and warps them to the output viewpoint by means of an affine transformation. Then, an image completion network is employed to seamlessly merge the warped patches and produce the final result. Park et al.~\cite{park2017transformation} draw inspiration from~\cite{zhou2016view} to relocate pixels visible both in the input and target view before using an image completion network based on adversarial training to refine the intermediate result.
\section{Model}
We present here the two fundamental stages of our approach, as illustrated in Fig.~\ref{modelfig}. In the first one \textit{(interpretable information extraction)}, we focus on acquiring high level interpretable information for each vehicle in the scene. That information is then exploited by the second stage (\textit{novel view completion}) to generate the final appearance of each vehicle individually. 

\subsection{Interpretable Information Extraction}
During this stage, high level interpretable information is gathered from raw RGB frames. Vehicles are first detected and their trajectories predicted. However, these trajectories are bound to the 2D plane, which is not sufficient to produce realistic movements (e.g. a car taking a turn). As such, we also detect vehicle 2D keypoints and align them to 3D ones by means of a perspective-n-point algorithm, obtaining a roto-translation matrix. This way, we can lift both the vehicle and the trajectory from 2D to 3D, and simulate realistic movements. 

Components from this stage are not the focus of this work. In fact, we're not interested in advancing the research in any of these tasks here, and we make use of pre-trained state-of-the-art methods when possible.

\subsubsection{Vehicle Detection}
We employ SSD detection network~\cite{liu2016ssd} to detect vehicles in the scene. Starting from the input frame, SSD outputs a set of bounding boxes (one for each detected object) in a single forward, along with their class probabilities. We filter the bounding boxes to keep only those associated with a vehicle, and use them to crop the visual appearance of each of them.

\subsubsection{Trajectory Prediction}
We employ TrackletNet~\cite{wang2019tracklet} as a trajectory predictor; it compares each vehicle tracklet -- composed by the detected bounding box and the appearance features -- along a time window of 64 consecutive frames. Using a similarity measure between tracklets, a graph is created where vertices under a certain distance threshold represent the same object.

\subsubsection{Keypoints Localisation}
We adapt a state-of-the-art network for human pose estimation, namely \textit{Stacked Hourglass}~\cite{newell2016stacked}, to localise vehicle keypoints. The network is characterised by a tunable number of encoder-decoder stacks. The final decoder outputs a set of planes (one per keypoint) where the maximum value localises the keypoint location. We change the final output structure to produce 12 keypoints: (i) four wheels, (ii) four lights, and (iii) four front and back windshield corners.

\subsubsection{Pose estimation}\label{subsub:pose} We frame the vehicle pose estimation as a \textit{perspective-n-point} problem, leveraging correspondences between 2D and 3D keypoints. While the former are the outputs of the previous step, the latter come from annotated 3D vehicle models. We exploit the 10 annotated models included in Pascal3D+, and we train a VGG19-based network~\cite{simonyan2014very} to predict the correspondent model given the vehicle crop. We argue these 10 CADs cover the vast majority of urban vehicles, as they have been deemed sufficient to annotate all vehicle images in the Pascal3D+ car set by authors from~\cite{xiang2014beyond}. Then, we adopt a \textit{Levenberg-Marquardt}~\cite{darcis2018poselab} iterative optimization strategy to find the best roto-translation parameters by minimizing the reprojection error or \textit{residual} between the 2D original keypoints and the correspondent 3D projections. We follow the stop criteria presented in~\cite{darcis2018poselab}. Once the source roto-translation matrix $V_s$ is known, the predicted model can follow the 3D lifted trajectory by applying consecutive transformations defined by the vehicle trajectory -- i.e. the roto-translation between consecutive trajectory positions converted from pixel to GPS meter coordinates. After each transformation we obtain the target roto-translation matrix $V_t$.

\subsection{Novel View Completion}
Once we know what to move and where to move it, we require a method to condition a reprojected 3D model with the original 2D appearance from the vehicle detection module. Theoretically, any view synthesis approach from Sec.~\ref{sec:related} can be used. In practice, a vast majority of them~\cite{jaderberg2015spatial, yang2015weakly, tatarchenko2016multi,park2017transformation} is only able to handle a specific setting known as \textit{"look-at-camera"}, where the vehicle is placed in the origin and the camera $z$ axis points at it. However, in our setting both $V_s$ and $V_t$ are generic roto-translation matrices. Moreover, some of the methods involve voxel spaces~\cite{jaderberg2015spatial}, which makes infeasible finding a correspondence.

Because our focus is on real-world data, we also exclude works which require direct training supervision and can thus only been trained on synthetic data~\cite{yang2015weakly}. In fact, as of today no real-world vehicles dataset can be exploited for supervised novel view synthesis training, as they all lack multiple views for the same vehicle annotated with pose information. This also prevent us from using any method based on~\cite{wang2018high} for this task. 
In the following, we thus employ two approaches~\cite{esser2018variational,palazzi2019warp} that are able to handle generic transformations and can be trained in an unsupervised fashion on real-world data.

Giving as input the crop depicting the vehicle $x_s$ observed by a source camera viewpoint $V_s$, we project the 3D model with the roto-translation outlined by $V_t$. To enrich the representation, we render a $2.5D$ sketch with normal information. The newly produced output is then pasted into a static background, and the process repeated for each moving vehicle.

We rely on foreground suppression to generate a static background for the output clip. We also experimented with inpainting networks~\cite{nazeri2019edgeconnect} but found that results were less realistic by visual inspection due to the presence of several artefacts. We leave further investigation and the extension to moving cameras as a future work.

\section{Experiments}
In this section we present, both visually and quantitatively, the results of our proposed pipeline and we compare them with those from various end-to-end approaches (referred to as baselines in the following). We also introduce the employed datasets and the metrics of interest, as well as implementation details to ensure experiments reproducibility.

\subsection{Datasets}

\subsubsection{CityFlow~\cite{tang2019cityflow}} is a multi-target multi-camera tracking and re-identification vehicle dataset, introduced for the 2019 Nvidia AI City Challenge. It comprises more than 3 hours of high resolution traffic cameras videos with more than $200K$ bounding boxes and $600$ vehicles identities, split between train and test sets. The dataset also includes homography matrices for bird's eye visualisation. Vehicles detection and tracking have been annotated automatically using SSD~\cite{liu2016ssd} and TrackletNet~\cite{wang2019tracklet} as detector and tracker respectively. All baselines have been trained on the train split of this dataset.

\subsubsection{Pascal3D+~\cite{xiang2014beyond}} is composed of 4081 training and 1024 testing images, preprocessed to guarantee the vehicle is completely visible. Every image is also classified into one of ten 3D models. Both 3D and 2D keypoints are included. Because 2D keypoints localisation is crucial in our pipeline, we extend the Pascal training set by including frames from CarFusion~\cite{dinesh2018carfusion}. We train models for our first stage on this dataset to ensure the generalisation of our approach when tested on CityFlow.

\subsection{Evaluation Metrics}
We evaluate all methods using both pixel level and perceptual metrics. The former evaluates the exact spatial correspondence between the predicted and the ground truth target and are very sensitive to 2D transformations (e.g. translations). Contrarily, the latter evaluates the matching between the content of the two images. 
It is worth noting that we only compare a tight crop around each vehicle for all methods, instead of the full generated image. We argue this choice leads to a better understanding of true performance, because it removes a vast portion of the image -- with only static background in it -- we are not interested in evaluating.

\subsubsection{Pixel wise Metrics}
We employ the Mean Squared Error (MSE) as a measure of pixel distance between the target crop $x_t$ and the predicted one $x_p$ as follows:
\begin{equation}
    MSE(x_t,x_p) = \|x_t - x_p\|_2^2
\end{equation}
values are then averaged over to compute the final score.

\subsubsection{Perceptual Metrics}
We employ the Structural Similarity Index (SSIM)~\cite{wang2004image} as a measure of the degradation in the image quality due to image data manipulation, defined as:
\begin{equation}
    SSIM(x_t, x_p) = \frac{(2\mu_{x_t}\mu_{x_p} + c_1)(2\sigma_{x_t x_p} + c_2)}{(\mu_{x_t}^{2} + \mu_{x_p}^{2} + c_1)(\sigma_{x_t}^{2} + \sigma_{x_p}^{2} + c_2)}
\end{equation}

\begin{figure*}
    \centering
    \includegraphics[width=0.94\textwidth]{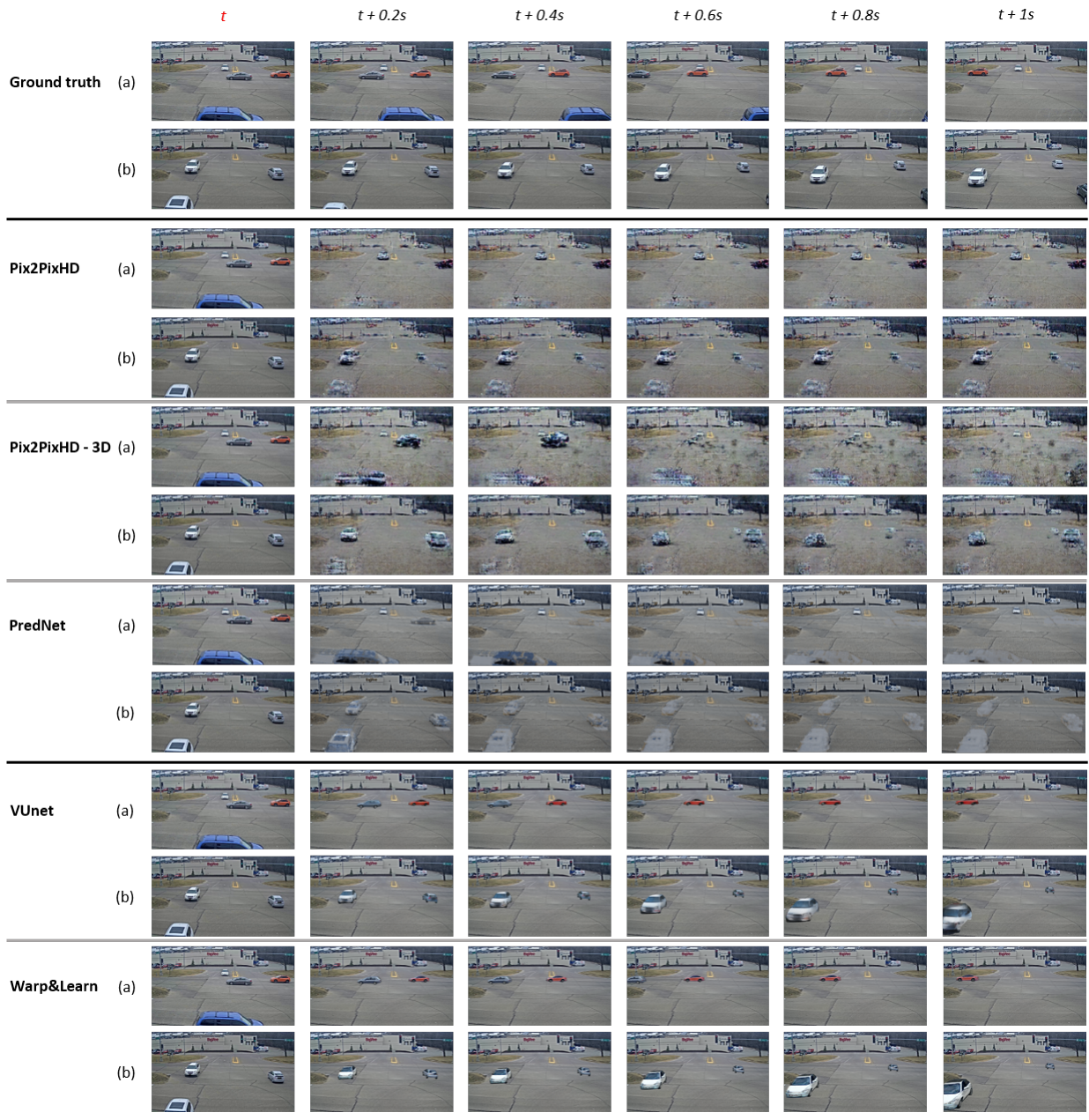}
    \caption{Visual results using the methods (Pix2PixHD, Pix2PixHD-3D, PredNet, VUnet, Warp\&Learn) on two ground truth video sequences (a) and (b) with different vehicles behaviour. Images at time \textit{t} refer to the ground truth, while images within 1 second in the future represent a method prediction.}
    \label{fig:visu_results}
\end{figure*}

As another measure of content similarity, we measure the Fréchet Inception Distance~(FID)~\cite{heusel2017gans,lucic2018gans} computed between activations from the last convolutional layer of an InceptionV3 model pretrained on ImageNet~\cite{krizhevsky2012imagenet}. We compute the FID as follow:
\begin{equation}
FID =\|m_t-m_p\|_2^2+  Tr \bigl(C_t + C_p-2\bigl(
C_t C_p\bigr)^{1/2}\bigr)
\end{equation}
where $m$, $C$ refer to the mean and covariance and follow the same notation as above for target and predicted image.

Finally, we also compute the Inception Score (IS)~\cite{salimans2016improved} to measure the generated images variety as:
\begin{equation}
    IS(G) = \textit{exp}(\frac{1}{N}\sum_{i = 1}^{N} D_{KL}(p(y \mid x^{(i)} \| \hat{p}(y)))
\end{equation}
where $x$ is an image, $N$ the total number of samples and $p(\hat{y})$ an empirical marginal class distribution.

\begin{figure*}[t]
    \centering
    \includegraphics[width=0.95\textwidth]{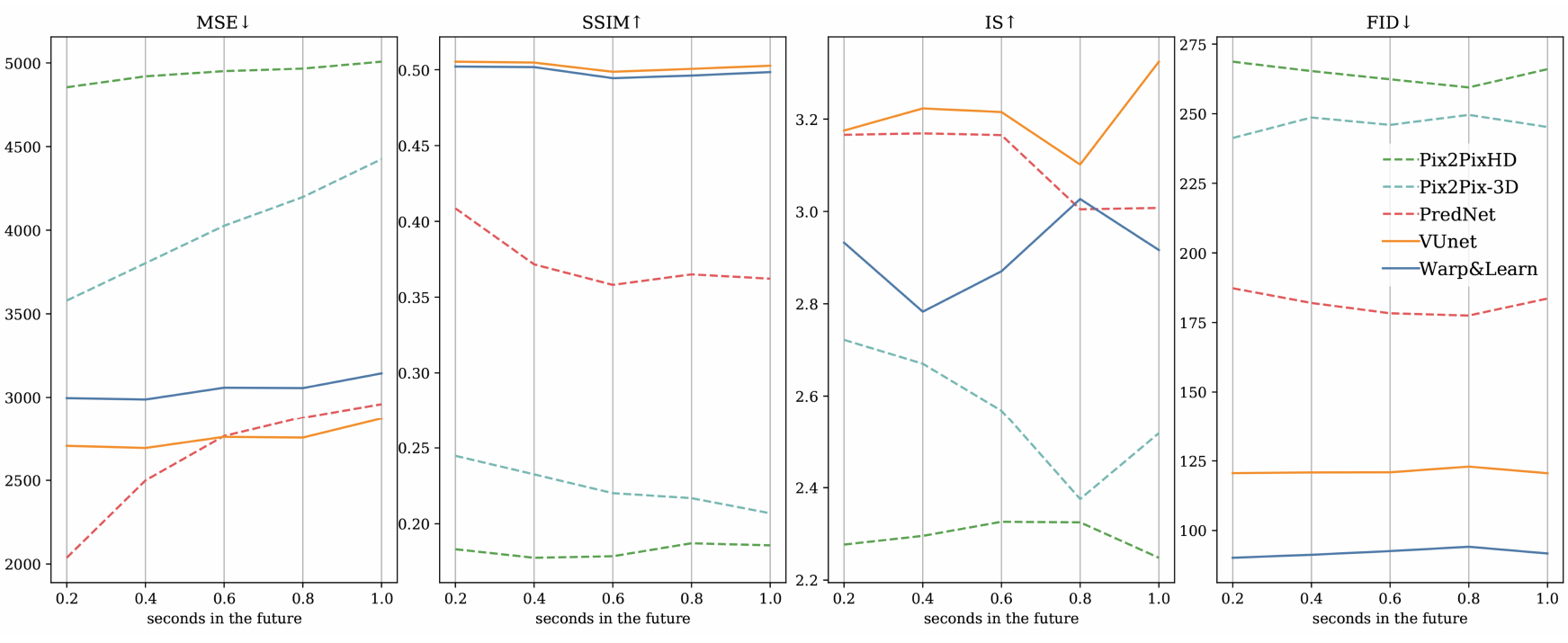}
    \caption{Comparison between our approach with two different types of image completion network (solid lines) and multiple baselines (dash lines) using Mean Squared Error (MSE)~(lower is better), Structural Similarity Index (SSIM)~(higher is better), Inception Score (IS)~(higher the better) and Frechet Inception Distance (FID)~(lower is better).}
    \label{fig:result_graph}
\end{figure*}

\begin{table}[t]
    \caption{Comparison on the test set using Mean Squared Error (MSE). Each column refers to a future displace. Lower is better.}
    \label{tab:mse}
    \centering
    \begin{tabular}{l|c|c|c|c|c}
        \hline
            Method & \textit{+0.2s} & \textit{+0.4s} & \textit{+0.6s} & \textit{+0.8s} & \textit{+1.0s}\\
        \hline
            Pix2PixHD~\cite{wang2018high} & 4854 & 4919 & 4950 & 4966 & 5007 \\
            Pix2Pix-3D  & 3579 & 3802 & 4026 & 4198 & 4424 \\
            PredNet~\cite{lotter2016deep}  & \textbf{2037} & \textbf{2499} & 2765 & 2877 & 2959 \\
        \hline
            Our(VUnet~\cite{esser2018variational})  & 2705 & 2692 & \textbf{2759} & \textbf{2755} & \textbf{2870} \\
            Our(Warp\&Learn~\cite{palazzi2019warp})  & 2996 & 2987 & 3058 & 3055 & 3153 \\
        \hline
    \end{tabular}
\end{table}

\begin{table}[t]
    \caption{Comparison on the test set using Structural Similarity Index (SSIM). Each column refers to a future displace. Higher is better.}
    \label{tab:ssim}
    \centering
    \begin{tabular}{l|c|c|c|c|c}
        \hline
            Method & \textit{+0.2s} & \textit{+0.4s} & \textit{+0.6s} & \textit{+0.8s} & \textit{+1.0s}\\
        \hline
            Pix2PixHD~\cite{wang2018high} & 0.18 & 0.17 & 0.17 & 0.18 & 0.18 \\
            Pix2Pix-3D  & 0.24 & 0.23 & 0.22 & 0.21 & 0.20 \\
            PredNet~\cite{lotter2016deep}  & 0.40 & 0.37 & 0.35 & 0.36 & 0.36 \\
        \hline
            Our(VUnet~\cite{esser2018variational})  & \textbf{0.50} & \textbf{0.50} & \textbf{0.49} & \textbf{0.50} & \textbf{0.50} \\
            Our(Warp\&Learn~\cite{palazzi2019warp})  & \textbf{0.50} & \textbf{0.50} & \textbf{0.49} & 0.49 & 0.49 \\
        \hline
    \end{tabular}
\end{table}

\begin{table}[t]
    \caption{Comparison on the test set using Inception Score (IS). Each column refers to a future displace. Higher is better}
    \label{tab:is}
    \centering
    \begin{tabular}{l|c|c|c|c|c}
        \hline
            Method & \textit{+0.2s} & \textit{+0.4s} & \textit{+0.6s} & \textit{+0.8s} & \textit{+1.0s}\\
        \hline
            Pix2PixHD~\cite{wang2018high} & 2.27 & 2.29 & 2.32 & 2.32 & 2.24 \\
            Pix2Pix-3D  & 2.72 & 2.67 & 2.56 & 2.37 & 2.51 \\
            PredNet~\cite{lotter2016deep}  & 3.16 & 3.16 & 3.16 & 3.00 & 3.00 \\
        \hline
            Our(VUnet~\cite{esser2018variational})  & \textbf{3.17} & \textbf{3.22} & \textbf{3.21} & \textbf{3.10} & \textbf{3.32} \\
            Our(Warp\&Learn~\cite{palazzi2019warp})  & 2.93 & 2.78 & 2.87 & 3.02 & 2.91 \\
        \hline
    \end{tabular}
\end{table}

\begin{table}[t]
    \caption{Comparison on the test set using Frechet Inception Distance (FID). Each column refers to a future displace. Lower is better}
    \label{tab:fid}
    \centering
    \begin{tabular}{l|c|c|c|c|c}
        \hline
            Method & \textit{+0.2s} & \textit{+0.4s} & \textit{+0.6s} & \textit{+0.8s} & \textit{+1.0s}\\
        \hline
            Pix2PixHD~\cite{wang2018high} & 274.2 & 268.6 & 265.3 & 262.3 & 259.4 \\
            Pix2Pix-3D  & 240.6 & 241.2 & 248.6 & 245.9 & 249.5 \\
            PredNet~\cite{lotter2016deep}  & 197.1 & 197.2 & 196.4 & 193.4 & 196.3 \\
        \hline
            Our(VUnet~\cite{esser2018variational})  & 192.8 & 187.3 & 182.0 & 178.3 & 177.49 \\
            Our(Warp\&Learn~\cite{palazzi2019warp})  & \textbf{90.4} & \textbf{90.22} & \textbf{91.2} & \textbf{92.6} & \textbf{94.1} \\
        \hline
    \end{tabular}
\end{table}

\subsection{Baselines}
\subsubsection{Pix2Pix} We adapt Pix2PixHD~\cite{wang2018high} for future frame prediction. Because it is trained in an end-to-end fashion, we can trivially include any high level information in the input. Still, we need to condition somehow the output to generate a specific frame (e.g. 0.2 seconds in the future) given the input image. As such, we stack the input with a set of binary maps along the channels dimension. During training, a random frame in the future is selected and the correspondent map is set to 1, while the others are set to 0. It is worth noting that predicting movements given a single input image is clearly an ill-posed task. For this reason, we also include another Pix2Pix baseline -- referred as Pix2Pix-3D in the following -- which is time aware. We provide this baseline with a set of past frame, and replace 2D convolution in the encoder with 3D ones. As such, this baseline version has access to past frames and can therefore exploit temporal information to determine if and how a vehicle is moving. However, this comes with an increase in memory footprint.

\subsubsection{PredNet} 
We also adapt PredNet~\cite{lotter2016deep} as a recurrent-based approach to the task. Differently from the previous two, this baseline generates frames in the future via a  recurrent structure. However, generated frames have to be forwarded as part of the input to produce frames further in the future, causing errors to propagate and performance to degrade in the long run.

\begin{figure*}
    \centering
    \includegraphics[width=0.91\textwidth]{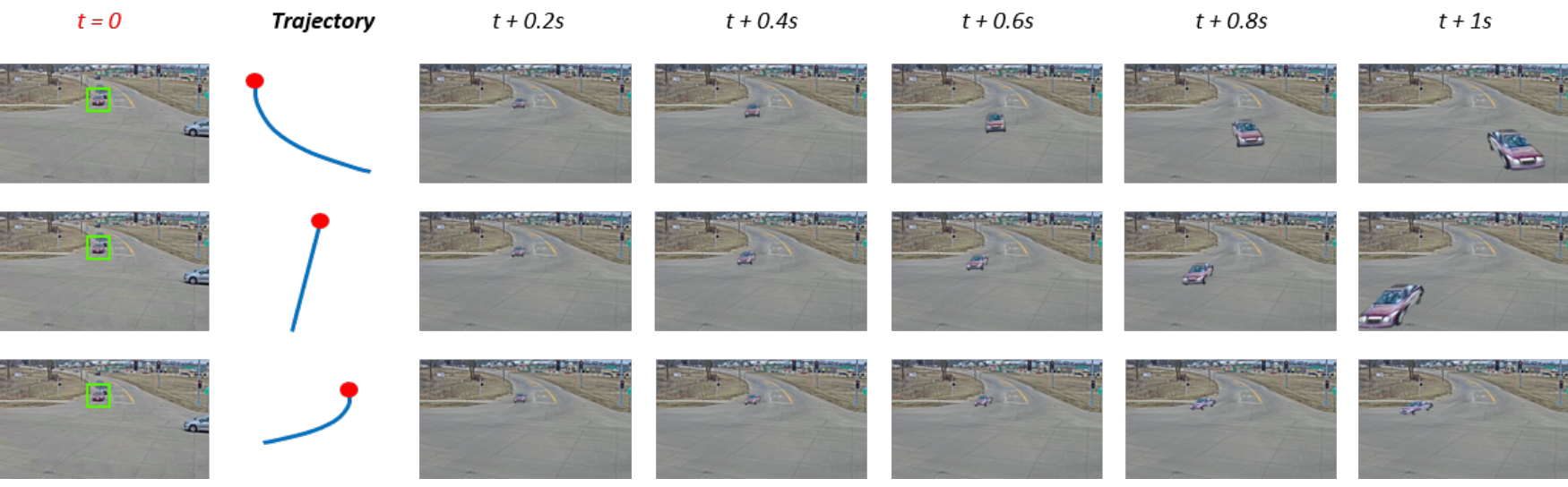}
    \caption{Visual results from our approach for three plausible futures constrained over different trajectories. The detected vehicle follows the indicated trajectory while preserving the original appearance. Best viewed in color.}
    \label{trajectoriesfig}
\end{figure*}

\subsection{Implementation Details}
All baselines are trained for 150 epochs on frames from the Cityflow train set, resized to 640x352 pixels. Pix2PixHD-based models employ batch size equal to 4 with initial learning rate of $2e^{-4}$ and linear decay as defined in~\cite{isola2017image}. PredNet is trained according to the original paper parameters.

\noindent As for our pipeline, the Keypoints localisation network is trained for 100 epochs employing batch size 10 and a learning rate of $1e^{-3}$ halved every 20 epochs, while VUnet and Warp\&Learn models are trained following the policies described respectively in~\cite{esser2018variational, palazzi2019warp}. All models except those for detection and tracking are trained on Pascal3D+ vehicle images resized to 256x256 pixels.

\noindent Code has been developed using the PyTorch~\cite{paszke2017automatic} framework and the Open3D library~\cite{zhou2018open3d} has been employed to manipulate and render the 3D CAD in the scene. Inference is performed on Cityflow test set videos resized to 1280x720 pixels. Code is available at \href{https://github.com/alexj94/future_urban_scene_generation}{https://github.com/alexj94/future\_urban\_scene\_generation}.

\subsection{Results}
Comparisons of the different methods are reported in Tables~\ref{tab:mse}, \ref{tab:ssim}, \ref{tab:is}, \ref{tab:fid} and in Figure~\ref{fig:result_graph}. Our proposed approach outperforms the baselines for all metrics in the long run, while scores second behind PredNet for the first two predictions according to the MSE. 

However, it is worth noting how Prednet is not capturing movements in an effective way, as shown in Figure~\ref{fig:visu_results}. While first outputs looks realistic, performance degrades quickly when predictions are employed as inputs for the LSTM. Our approach proves to be superior for all the metrics that reward the content realism (i.e. FID, IS and SSIM) and to suffer less performance degradation for long time predictions. This highlights how focusing on individual vehicles is crucial in visual future scene prediction. Between the two novel view synthesis methods, \cite{esser2018variational} achieves better performance for 2 (IS, MSE) out of 4 metrics, with comparable results for the SSIM. On the other hand \cite{palazzi2019warp} outperforms all other methods by a consistent margin for the FID metric.

Figure~\ref{fig:visu_results} reports a visual comparison between different methods on two ground truth sequences. It can be appreciated how our approach produces higher quality results, both for the static background and the foreground. On the other hand, baseline methods struggle to produce crisp images, often resulting in extremely blurry images. As expected, Pix2PixHD fails completely to predict vehicles movement and collapses into a static image output. While Pix2Pix-3D partially solves this issue, it still focus mostly on the background. Finally, PredNet is able to guess correctly the evolution of the scene, but performance degrade in the long run, with vehicles progressively fading away. It is worth noting that for the first sequence the vehicle closer to the camera is not modelled by our method, and thus disappears immediately. This is due to an SSD miss-detection. Even though our final output depends on many modules we argue this is not a weakness in the long run. In fact, it's trivial to replace a single component with another with better performance, while the same consideration does not hold true for end-to-end approaches.

\subsection{Constrained Futures Generation}
Thanks to its two-stages pipeline our methods can be trivially constrained using high-level interpretable information. Figure~\ref{trajectoriesfig} illustrate an example of this process where the constraint is provided in the form of trajectories. Starting from the same input frame, three futures are generated by providing different trajectories. It can be appreciated how the vehicle closely follows the designated path, which can be easily drawn by a non-expert user. Other constrains that can be provided out-of-the-box include a different CAD model or a different appearance. The same interaction is not well-defined for end-to-end methods.
\section{Conclusions}
In this work we presented a novel pipeline for predicting the visual future appearance of an urban scene. We propose a novel approach as an alternative to end-to-end solutions, where human interpretable information is included into the loop and every actor is modelled independently. Existing state-of-the-art methods or the user can both be sources for that information. Furthermore, the final visual output is conditioned onto that by design. We demonstrate the performance superiority of our pipeline with respect to traditional end-to-end baselines through an extensive experimental section. Moreover, as shown in Fig.~\ref{trajectoriesfig}, we visually illustrate how our method can generate diverse realistic futures starting from the same input by varying the provided interpretable information.

\section*{Acknowledgements}
This research was supported by MIUR PRIN project "PREVUE:
PRediction of activities and Events by Vision in an Urban Environment", grant ID E94I19000650001.


%
\IEEEpeerreviewmaketitle

\bibliographystyle{IEEEtran}
\bibliography{main.bib}
%

\end{document}